\documentclass[11pt,a4paper]{article}
\usepackage[hyperref]{emnlp2020}
\usepackage{times}
\usepackage{latexsym}

\usepackage{microtype}

\aclfinalcopy 

\usepackage{bm}
\usepackage{booktabs}
\usepackage{array}
\usepackage{multirow}
\usepackage{makecell}
\usepackage{arydshln}
\usepackage{graphicx}
\usepackage{color}
\usepackage{amsmath}
\usepackage{amssymb}
\usepackage{tikz}
\newcommand*{\Scale}[2][4]{\scalebox{#1}{$#2$}}  
\usepackage[noend]{algpseudocode}
\usepackage[linesnumbered, ruled, vlined]{algorithm2e}

\title{FewJoint: A Few-shot Learning Benchmark for Joint Language Understanding}

\author{
	Yutai Hou$^{1}$,
	Jiafeng Mao$^{1}$,
	Yongkui Lai$^{1}$,
	Cheng Chen$^{1}$, \\
	\textbf{Wanxiang Che}$^{1}$\thanks{$\ $ Corresponding author.}~,
	\textbf{Zhigang Chen}$^{2}$,
	\textbf{Ting Liu}$^{1}$
	\\
	$^{1}$Research Center for Social Computing and Information Retrieval, \\ Harbin Institute of Technology \\
	$^{2}$State Key Laboratory of Cognitive Intelligence, Hefei, China  \\
	{\tt \{ythou, jfmao, klai, car, tliu\}@ir.hit.edu.cn,} \\ 
	{\tt 170400202@stu.hit.edu.cn,} {\tt zgchen@iflytek.com} \\
}

\date{}

\begin{document}
\maketitle
\begin{abstract}
Few-shot learning (FSL) is one of the key future steps in machine learning and has raised a lot of attention.
However, in contrast to the rapid development in other domains, such as Computer Vision, the progress of FSL in Nature Language Processing (NLP) is much slower. 
One of the key reasons for this is the lacking of public benchmarks. 
NLP FSL researches always report new results on their own constructed few-shot datasets, which is pretty inefficient in results comparison and thus impedes cumulative progress.
In this paper, we present FewJoint, a novel Few-Shot Learning benchmark for NLP. 
Different from most NLP FSL research that only focus on simple N-classification problems, our benchmark introduces few-shot joint dialogue language understanding, which additionally covers the structure prediction and multi-task reliance problems. 
This allows our benchmark to reflect the real-word NLP complexity beyond simple N-classification. 
Our benchmark is used in the few-shot learning contest of SMP2020-ECDT task-1.\footnote{The Eighth China National Conference on Social Media Processing. Link: \url{https://smp2020.aconf.cn/smp.html}} 
We also provide a compatible FSL platform to ease experiment set-up.\footnote{The dataset and platform is available at \url{https://github.com/AtmaHou/MetaDialog}}

\end{abstract}

\section{Introduction}

Deep learning has achieved significant successes, but these successes heavily rely on massive annotated data.
Few-Shot Learning (FSL) is one of the keys to breaking such shackle, and commits to learning new tasks with only a few examples (usually only one or two per category) \cite{miller2000learning,fei2006one,lake2015human,matching}.
FSL has made impressive progress in many areas, such as computer vision (CV) \cite{matching,prototypical,yoon2019tapnet}.
But the progress of FSL in natural language processing (NLP) is much slower. 
One of the primary constraints is the lack of a unified benchmark for few-shot NLP, thus new methods cannot be easily compared and iteratively improved.

Existing few-shot NLP researches mainly focus on simple N-classification problems, such as 
text classification \cite{textSun2019hierarchical,textGeng2019induction,yan2018few,yu2018diverse,DBLP:conf/iclr/BaoWCB20,policyVlasov2018few} and 
entity relation classification \cite{relationLv2019adapting,relationGao2019neural,relationYeL19}. 
However, on one hand, these works often report results on their own constructed few-shot data, which is pretty inefficient in results comparison and thus hinders cumulative progress.
On the other hand, these simple N-classification problems cannot reflect the complexity of real-world NLP tasks. 
NLP tasks often face the challenges of structure prediction problems, such as sequence labeling \cite{CRF,hou2020fewshot} and parsing \cite{klein2003accurate}.
More importantly, different NLP tasks are often deeply related to each other, i.e. multi-task problems \cite{worsham2020multi}. 
One typical scenario of complex NLP is the Dialogue Language Understanding problem, which includes two sub-tasks: Intent Detection (text classification) and Slot Tagging (sequence labeling). 
As a multi-task problem, these two sub-tasks are proved to strongly promote and depend on each other \cite{chen2019bert,goo2018slot}.

One of the main obstacles in constructing the NLP FSL benchmark comes from the special evaluation paradigm of FSL.
Few-shot models are usually first pre-trained on data-rich domains (to learn general prior experience) and then tested on unseen few-shot domains.
Thus, FSL evaluations always need a lot of different domains to conquer the result-randomness from domain selection and limited learning shots. 
But it is often hard to gather enough domains for NLP tasks. 
To solve this, existing works \cite{DBLP:conf/iclr/BaoWCB20,relationGao2019fewrel} construct fake domains from a single dataset.
They split all labels into training labels and testing labels. 
Then, they construct fake pre-training and testing domains with training and testing labels respectively, so that testing labels are unseen during pre-training.
Such simulation can yield plenty of related domains, but lacks reality and only works when the label set is large. 
Actually, splitting labels is impractical for many real-world NLP problems. 
For example of the Name Entity Recognition, the label sets are often too small to split (usually 3 or 5 labels).

\begin{figure*}[t]
	\centering
	\begin{tikzpicture}
	\draw (0,0 ) node[inner sep=0] {\includegraphics[width=2.08\columnwidth, trim={4cm 2.5cm 2cm 3cm}, clip]{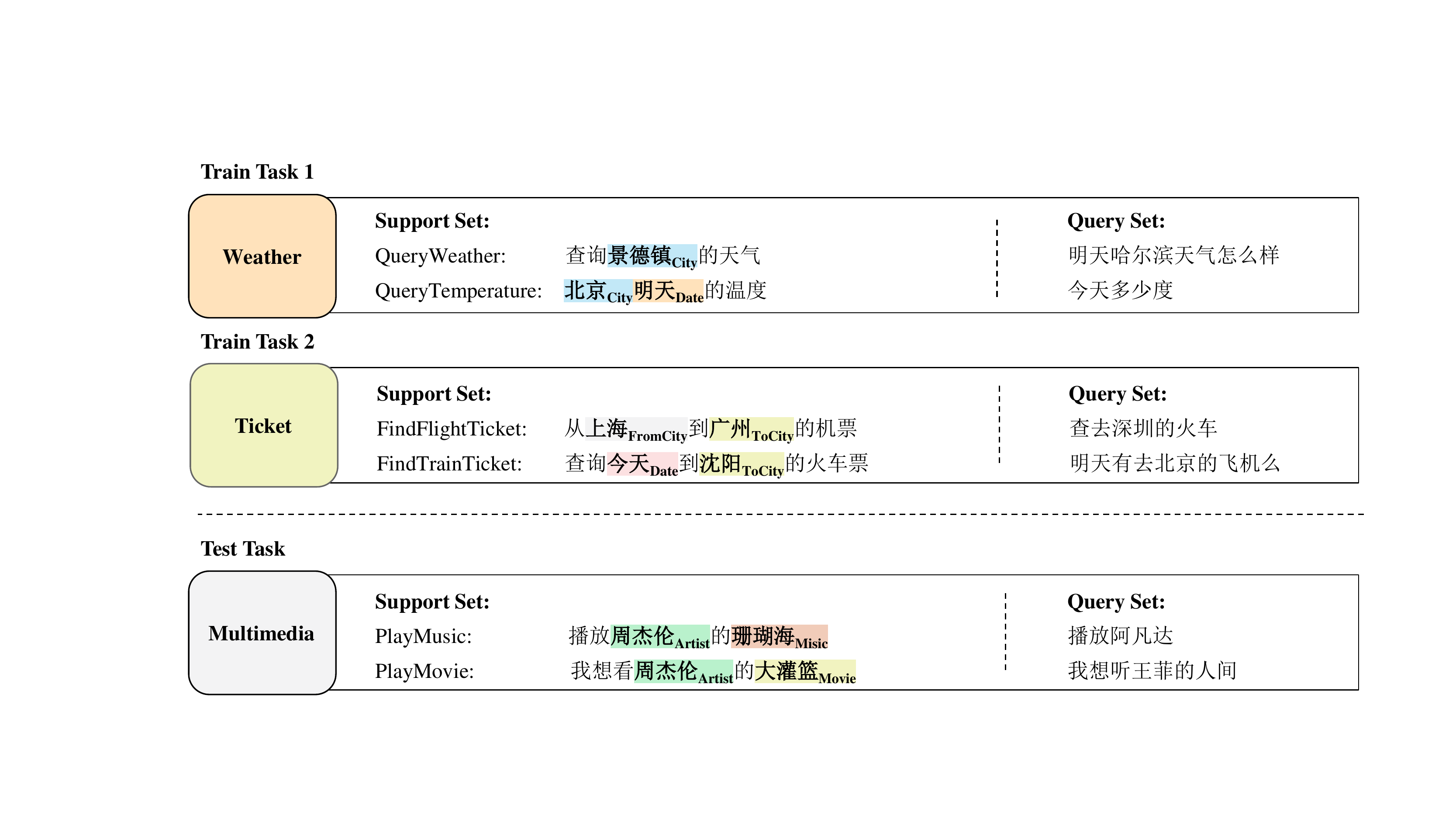}};
	\end{tikzpicture}
	\caption{\footnotesize
		Training and testing examples of the few-shot dialogue language understanding benchmark.
		We training the model on a set of source domains, and testing it on an unseen domain with only a support set. }\label{fig:train_test}
\end{figure*}

In this paper, we present FewJoint, a novel FSL benchmark for joint multi-task learning, to promote FSL research of the NLP area. 
To reflect the real word NLP complexities beyond simple N-classification, we adopt a sophisticated and important NLP problem for the benchmark: Task-oriented Dialogue Language Understanding. 
Task-oriented Dialogue is a rising research area that develops dialogue systems to help users to achieve goals, such as booking tickets.
Language Understanding is a fundamental module of Task-oriented Dialogue that extracts semantic frames from user utterances (See Figure \ref{fig:train_test}).
It contains two sub-tasks: Intent Detection and Slot Tagging. 
With the Slot Tagging task, our benchmark covers one of the most common structure prediction problems: sequence labeling.
Besides, thanks to the natural dependency between Intent Detection and Slot Tagging, our benchmark can embody the multi-task challenge of NLP problems.
To conquer randomness and make an adequate evaluation, 
we include 59 different dialogue domains from real industrial API, which is a considerable domain amount compared to all existing few-shot and dialogue data.
We also provide a Few-shot Learning platform to ease the experiment set up and comparison. 

In summary, our contribution is three-fold:
(1) We present a novel Few-shot learning benchmark with 59 real-world domains, which allows evaluating few-shot models without constructing fake domains.
(2) We propose to reflect real-world NLP complexities by covering the structure prediction problems and multi-task learning problems.
(3) We propose a Few-shot Learning platform to ease comparison and implement of few-shot methods.

\section{Problem Definition}\label{sec:p_def}
Before introducing the dataset, we present the definition of the few-shot language understanding problem here.

Starting from the notions, we define utterance $\bm{x} = (x_1, x_2, \ldots, x_n)$ as a sequence of words 
and define corresponding semantic frame as $\bm{y}=(c, \bm{s})$. 
$c$ is the intent label of the utterance and $\bm{s}$ is the slot label sequence of the utterance and defined as $\bm{s} = (s_1, s_2, \ldots, s_n)$.
A domain $\mathcal{D}  = \left\{(\bm{x}^{(i)},\bm{y}^{(i)})\right\}_{i=1}^{N_D}$ is a set of $(\bm{x},\bm{y})$ pairs. 
For each domain, there is a corresponding domain-specific label set $\mathcal{L_D}$.
To simplify the description and ease understanding, 
we combine the label set definition of intent and slot, and assume that the number of labels $N$ is the same for all domains.

In few-shot learning scenarios, models are usually first trained on a set of source domains $\left\{\mathcal{D}_1, \mathcal{D}_2, \ldots \right\}$, 
then evaluated on another set of unseen target domains $\left\{\mathcal{D}_1', \mathcal{D}_2', \ldots \right\}$. 
A target domain $\mathcal{D}_j'$ only contains few labeled examples,
which is called support set $\mathcal{S} = \left\{(\bm{x}^{(i)},\bm{y}^{(i)})\right\}_{i=1}^{N_\mathcal{S}}$.
$\mathcal{S}$ usually includes $K$ examples (K-shot) for each of $N$ labels (N-way).
Figure \ref{fig:train_test} shows an example of the training and testing process of 1-shot dialogue understanding.

The K-shot dialogue understanding task is then defined as follows: 
given an input query utterance $\bm{x} = (x_1, x_2, \ldots, x_n)$ and a K-shot support set $\mathcal{S}$ as references, 
find the most appropriate semantic frame $\bm{y}^*$ of  $\bm{x}$:
\[
\Scale[0.85]{
	\bm{y}^* = (y_1, y_2, \ldots, y_n) = \mathop{\arg\max}_{\bm{y}} \ \ p(\bm{y}  \mid  \bm{x}, \mathcal{S})
}.
\]

\section{Data Construction}
In this section, we introduce the construction process of the FewJoint dataset, which generally contains two steps.
Firstly, we collect and annotate a complete dialogue understanding corpus 59 domains (Section \ref{sec:collect} and Section \ref{sec:annotate}). 
Then, we split the corpus into training and unseen few-shot domains. We also sample support and query set to simulate few-shot scenarios (Section \ref{sec:simulate}).  

\subsection{Dialogue Collection}\label{sec:collect}
We collect dialogue utterance of real dialogue domains from the AIUI open dialogue platform of iFlytek.\footnote{\url{http://aiui.xfyun.cn/index-aiui}}
Before utterance collection, we select popular domains based on the frequency of API calls, such as ``Search information of corona-virus''. 
We ignore the domains that have no Intent or Slot Schema to ensure joint learning.
For schema definition, we leverage the semantic-frames and domains defined by AIUI, and also refine parts of domains to remove ambiguous labels.
Finally, we gather 59 different dialogue domains together with semantic frame definitions.

Given the semantic-frame definitions, we collect user utterances generally from two sources:
\begin{itemize}
	\item[(1)] Real user utterances.
	\item[(2)] Utterances written by the worker.
\end{itemize}
For the source (1), we sample existing user utterances from the AIUI platform and remove the sensitive information. 
For the source (2), 
four workers were asked to impersonate users of dialogue agents and write query utterances for the specific domains, such as querying weather.
The average ratio of utterances between source (1) and (2) is about 3 : 7.

\subsection{Data Annotation}\label{sec:annotate}
After collecting raw user utterances, we label each utterance with both intent (sentence level) and slot labels (token level).
The support sets in Figure \ref{fig:train_test} show examples of utterances with intent and slots annotations.

The annotating process consists of two steps: 
Firstly, we obtain rough annotation for each utterance by predicting semantic-frame with the testing-tools of the AIUI platform.
Then, the human workers validate each roughly annotated utterance and re-annotated the inappropriate ones.
The data was divided equally into four parts and then annotated by four workers respectively.
The four workers who participated in utterance writing are also responsible for this part.

After annotating, we perform data re-checking. 
Another three workers each checked all the data, and incorrect data are re-annotated.

\subsection{Simulation of Few-shot Scenarios}\label{sec:simulate}
Till now, we have collected the annotated dialogue corpus. 
To test the learning performance of few-shot learning models, we need to reconstruct the data into Few-Shot Learning (FSL) setting. 
In FSL setting, models are first pre-trained on data-rich domains and then tested on unseen domains with only a few-shot support set.
Figure \ref{fig:train_test} show an example of few-shot learning setting.

\paragraph{Few-shot Data Construction.}
To achieve the few-shot problem setting described in Section \ref{sec:p_def}, 
we reserve some domains as few-shot testing domains, which are unseen during training.
Specifically, we first split the 59 domains into three parts with no intersection: train, dev and test.
Then on each dev or test domains, we construct a $K$-shot support set and use the other data as the query set. 
Thus, FewJoint can simulate few-shot scenarios on the unseen testing domains: let the models predict the labels of the query samples with only few support examples.
Table \ref{tbl:domain} shows the details of domain division.

\begin{table*}[t]
	\centering
	\footnotesize
\begin{tabular}{p{1 \columnwidth}p{0.35 \columnwidth}p{0.35 \columnwidth}}
	\toprule
	 \multicolumn{1}{c}{\textbf{Train Domains (45)}} & \multicolumn{1}{c}{\textbf{Dev Domains (5)}} & \multicolumn{1}{c}{\textbf{Test Domains (9)}} \\
	\cmidrule(lr){2-2}
	\cmidrule(lr){3-3}
	\cmidrule(lr){1-1}
	queryCapital, app, epg, petrolPrice, dream, animalCries, historyToday, translation, sentenceMaking, carNumber, poetry, familyNames, match, clock, weightScaler, cityOfPro, airControl, website, stock, riddle, map, cookbook, music, calendar, crossTalk, wordsMeaning, new, health, home, video, telephone, weather, tvchannel, lottery, stroke, radio, contacts, bus, message, train, novel, email, cinemas, flight, childClassics
	& wordFinding, garbageClassify, holiday, joke, temperature
	& idiomsDict, timesTable, virusSearch, captialInfo, constellation, drama, length, story, chineseZodiac
	\\	
	\bottomrule
\end{tabular}
	\caption{The domains of FewJoint benchmark. }\label{tbl:domain}
\end{table*}

\paragraph{Reconstructing Testing Domains}
We reconstruct each test/dev domains into two part: a few-shot support set and a query set.
Here, a $K$-shot support set is manually constructed by the following principles:
\begin{itemize}
	\item Ensure each class (intent and slot) appeared at least $k$ times, while keeping the support set as small as possible. 
	\item Avoid duplication between support set and query set.
	\item Encourage diversity of both expressions and slot values of support set.
\end{itemize}

\paragraph{Reconstructing Training Domains}
The training set consists of 45 training domains, which provide prior experience to help quick learning on unseen domains.
For few-shot learning, there are two popular strategies for learning such prior experience and their data format is very different.
In our benchmark, we provide two kinds of training set format to support these two learning strategies \footnote{Benchmark users are free to re-construct training set into any format.}:

\begin{itemize}
	\item[(1)] Learn the feature encoding layer on all training data, which simply needs to combine all train domain utterances into a single pre-training set.
	\item[(2)] Learn the ability of learning quickly when given only a few examples, i.e. meta learning. This requires to reconstruct training set into a series of few-shot episodes (i.e. support set and query set pair).
\end{itemize}



The strategy (1) does not require special data processing.
To support the strategy (2), we need to sample query and support sets to construct few-shot learning episodes within training domains. 
Here, we adopt the \texttt{Minimum-including Algorithm} \cite{hou2020fewshot} to achieve automatic sampling of plentiful few-shot episodes.

\texttt{Minimum-including Algorithm} helps to sample support set for sequence labeling problems and multi-label problems, where a single instance may be associated with multiple labels. 
In these problem settings, the normal \textit{N-way K-shot} support set definition is inapplicable. 
Because different labels randomly co-occur in one sentence, and we cannot guarantee that each label appears $K$ times.
Take the dialogue understanding problem as an example, each utterance instance is often associated with multiple labels, including one intent and multiple slots.
For example in Figure \ref{fig:train_test}, in the 1-shot support set, the slot of ``FromCity'' appeared twice to ensure all labels appear at least once. 
Thus, Minimum-including Algorithm approximately builds K-shot support set $\mathcal{S}$ following two criteria: 

\begin{itemize}
	\item[(1)] All labels within the domain appear at least $K$ times in support set $\mathcal{S}$.
	\item[(2)] At least one label will appear less than $K$ times in $\mathcal{S}$ if any $(\bm{x},\bm{y})$ pair is removed from it. 
\end{itemize}

Algorithm \ref{algorithm} shows the detailed process.
%

\begin{algorithm}[t]
	\caption{Minimum-including}\label{algorithm}
	\footnotesize
	\begin{algorithmic}[1]
		\Require \# of shot $K$, domain $\mathcal{D}$, label set $\mathcal{L_D}$ \\
		Initialize support set $\mathcal{S}=\left\{ \right\}$, $\text{Count}_{\ell_j} = 0 $ 
		$(\forall \ell_j \in \mathcal{L_D})$
		\\
		
		\For{$\ell$ in $\mathcal{L_D}$} {
			\While{$\text{Count}_{\ell} < K $ }
			{From $\mathcal{D} \setminus \mathcal{S}$, randomly sample a $(\bm{x}^{(i)},\bm{y}^{(i)})$ pair that $\bm{y}^{(i)}$ includes $\ell$
				
				Add $(\bm{x}^{(i)},\bm{y}^{(i)})$ to $\mathcal{S}$
				
				Update all $\text{Count}_{\ell_j}$ 
				$(\forall \ell_j \in \mathcal{L_D})$
			}
		} \\
		
		\For{each $(\bm{x}^{(i)},\bm{y}^{(i)})$ in $\mathcal{S}$}
		{   
			Remove $(\bm{x}^{(i)},\bm{y}^{(i)})$ from $\mathcal{S}$ 
			
			Update all $\text{Count}_{\ell_j}$ 
			$(\forall \ell_j \in \mathcal{L_D})$ 
			
			\If{any $\text{Count}_{\ell_j} < $  K}
			{Put $(\bm{x}^{(i)},\bm{y}^{(i)})$ back to $\mathcal{S}$
				
				Update all $\text{Count}_{\ell_j}$ 
				$(\forall \ell_j \in \mathcal{L_D})$
			}
		}\\
		
		Return $\mathcal{S}$ 
		
	\end{algorithmic}
\end{algorithm}

\section{Statistic}
This section presents a detailed statistic for the constructed dataset. 
The statistic info of annotated raw paper is shown in Table \ref{tbl:stat_raw_data}
There are 6,694 utterances included in the corpus and the average length of utterance is 9.9 (number of Chinese characters). 
As mention before, we collect data for 59 real dialogue domains.
Among them, we reserve 14 domains as unseen few-shot domains for evaluation and use all the other 45 domains as training domains.
For evaluation, we select 9 domains as the test set, and use 5 for development.
Overall, our data set contains 143 different intents and 205 different slots.

\begin{table}[t]
	\centering
	\footnotesize
	\begin{tabular}{  l  c  }
	\toprule
	Item & Count \\
	\midrule
	Total number of utterances & 6,694 \\ 
	Average length of utterance & 9.9 \\
	
	\midrule
	Total number of domain & 59 \\
	Number of domain (train) & 45 \\
	Number of domain (dev) & 5 \\ 
	Number of domain (test) & 9 \\
	
	\midrule
	Total number of intents & 143 \\ 
	Average intents per. domain & 2.42 \\ 
	
	\midrule
	Total number of slots & 205 \\ 
	Average slots per. domain & 3.47 \\ 
	\bottomrule
\end{tabular}
\caption{
\footnotesize
	Statistic of raw data.
}\label{tbl:stat_raw_data}
\end{table}

Table \ref{tbl:stat_fs_data} shows the statistics of constructed few-shot data. 
The main setting (Used in the SMP2020 contest\footnote{The Evaluation of Chinese Human-Computer Dialogue Technology, SMP2020-ECDT task-1. Link: \url{https://smp2020.aconf.cn/smp.html}  }) is 3-shot language understanding, and we also provide 1, 5, 10 shots setting for extensive evaluation.
Support set size and query set size information of different shot setting is included in Table \ref{tbl:stat_fs_data}.
Besides, we also present the number of occurrences for each intent and slot in the support set, which satisfies our construction requirements for shots.

\begin{table}[t]
	\centering
	\footnotesize
	\begin{tabular}{ l  l  cc }
			\toprule
			Setting & & Support Sent. & Query Sent. \\
			\midrule
			\multirow{2}{*}{1-shot} & dev & 22 & 556 \\
			                        & test & 55 & 1,068  \\
			                        
			\multirow{2}{*}{3-shot} & dev & 66 & 511 \\
			                        &  test & 147 & 1,061 \\
			\multirow{2}{*}{5-shot} & dev & 108 & 470 \\
			                        & test & 233 & 969 \\
			\multirow{2}{*}{10-shot} & dev & 192 & 389 \\
			                        & test & 476 & 731 \\
			\midrule
			& & Ave. Intent in $S$ & Ave. Slot in $S$ \\
			\midrule
			\multirow{2}{*}{1-shot} & dev & 1.4 & 1.7 \\ 
			       & test & 1.3 & 2.0 \\ 
			\multirow{2}{*}{3-shot} & dev & 4.1 & 4.1 \\ 
			       & test & 3.7 & 4.6 \\ 
			\multirow{2}{*}{5-shot} & dev & 6.8 & 6.5 \\ 
			       & test & 6.0 & 7.3 \\ 
			\multirow{2}{*}{10-shot} & dev & 12 & 11.4 \\ 
			       & test & 11.7 & 15.0 \\ 
			\bottomrule
	\end{tabular}
	\caption{
		\footnotesize
		Statistic of few-shot benchmark.
	}\label{tbl:stat_fs_data}
\end{table}


\section{Experiments}
Following the setting of the SMP2020 contest, we provide baseline results for 3-shot.
During the experiments, we transfer the learned knowledge from source domains (training) to unseen target domains (testing) containing only a 3-shot support set.  
Three baseline models are evaluated: \textit{SepProto}, \textit{JointProto} and \textit{JointProto + Finetune}.

\subsection{Settings}

\paragraph{Evaluation}
To conduct a robust evaluation under few-shot setting, 
we validate the models on different domains and take the average score as final results.

There are three main metrics for evaluation: Intent Accuracy, Slot F1-score, Sentence Accuracy.
Specifically, we calculate the Slot F1-score on query samples with \texttt{conlleval} script.\footnote{\url{https://www.clips.uantwerpen.be/conll2000/chunking/conlleval.txt}}
For Sentence Accuracy, we consider that one sentence is correct only when all its slots and intent are correct, and vice versa.
All models are evaluated on the same support-query pairs for fairness. 

To control the nondeterministic of neural network training \citep{reimers-gurevych:2017:EMNLP2017}, 
we report the average score of 5 random seeds for all results.

\paragraph{Implements}
For sentence embedding, we average the token embedding provided by pretrained language model and we use uncased \textit{BERT-Base} \citep{BERT} here.
Also, we adopt embedding tricks of Pairs-Wise Embedding \cite{hou2020fewshot} and Gradual Unfreezing \cite{Howard2018UniversalLM}.
We use ADAM \citep{DBLP:journals/corr/KingmaB14} to train the models with batch size 4. 
The learning rate is set as 1e-5 for baseline models. 

During experiment, all baseline models are implemented with the provided few-shot platform.\footnote{\url{https://github.com/AtmaHou/MetaDialog}}

\subsection{Results}\label{sec:main_res}
Here, we evaluate the two types of few-shot learning strategies on 3-shot dialogue language understanding: 
(1) non-fine-tune based methods (\textit{SepProto},\textit{JointProto}) and (2)  fine-tune based methods (\textit{JointProto + Finetune}).
All the baseline models are provided by our few-shot learning platform:


\paragraph{SepProto} is a similarity metric based few-shot learning model with a prototypical network \citep{prototypical}.
It first averages the embeddings of each label's support examples as prototypes, and compares the similarity distance between the query instance and each prototype under a certain distance metric. 
Then it classifies an item to its closest labels.
During the experiment, it is pre-trained on source domains and then directly works on target domains without fine-tuning. 

\paragraph{JointProto} is also a prototypical model similar to SepProto. 
The difference is two-fold: (1) we adopt the logits-dependency trick proposed by \cite{goo2018slot} to achieve joint learning of intent detection and slot filling.
(2) intent detection and slot filling share same BERT embedding.
Similar to SeqProto, it is pre-trained on source domains and then directly works on target domains without fine-tuning.

\paragraph{JointProto + Finetune} is a joint language understanding model same to JointProto. 
Its difference from JointProto is that it is further finetuned on the support set of target domains.

The evaluation results are showed in Table \ref{tbl:main}.
As the results show, JointProto outperforms SepProto on both intent detection and slot filling, which indicates that additional information from joint-learning tasks can improve performance. 
For the Sentence Acc., JointProto significantly outperforms SepProto for 7.25 percent. 
This demonstrates the overall superiority of joint few-shot language understanding comparing to vanilla few-shot methods.
The improvements mainly come from two aspects: (1) Another task's logits serves as additional evidence for prediction. (2) Embedding sharing introduces additional supervisory signals to achieve better BERT finetuning. 

When comparing JointProto + Finetune and JointProto, huge improvements are witnessed on Sentence Acc and Slot F1. 
This shows on 3-shot setting, finetuning can further boosts the few-shot language understanding performance by providing domain specific knowledge.

\begin{table}[t]
	\centering
	\footnotesize
	\renewcommand\arraystretch{1.2}
	\resizebox{\linewidth}{!}{
		\begin{tabular}{lccc}
			\toprule
			\textbf{{Models}} & \multicolumn{1}{c}{\textbf{Intent Acc.}} & \multicolumn{1}{c}{\textbf{Slot F1}} & \multicolumn{1}{c}{\textbf{Sentence Acc.}} \\
			\midrule
			\textbf{SepProto}        & 72.30 & 34.11 & 16.40 \\
			\textbf{JointProto}      & \textbf{78.46} & 40.37 & 23.65 \\
			\textbf{JointProto + Finetune} & 73.82 & \textbf{61.79} & \textbf{38.97} \\
			\bottomrule
		\end{tabular}
	}
	\caption{
		\footnotesize
		Main result of baselines.
	}\label{tbl:main}
	
\end{table}

\section{Conclusion}

In this paper, we present a novel few-shot learning benchmark for NLP tasks, which is the first few-shot NLP benchmark for joint multi-task learning. 
Compared to existing few-shot learning data, our benchmark reflects real-world NLP complexities better by covering the structure prediction problem and multi-task learning problem. 
Also, our benchmark consists of 59 real dialogue domains.
This allows to evaluate few-shot model without constructing fake domain like existing works.

\bibliography{smp_data}
\bibliographystyle{acl_natbib}

\end{document}